\documentclass{article}

\usepackage[accepted]{eiml_icml2026}

\usepackage[T1]{fontenc}
\usepackage[utf8]{inputenc}
\usepackage{microtype}
\usepackage{amsmath,amssymb,mathtools}
\usepackage{booktabs}
\usepackage{tabularx}
\usepackage{multirow}
\usepackage{graphicx}
\usepackage{xcolor}
\usepackage{enumitem}
\usepackage{url}
\usepackage[colorlinks=True,linkcolor=orange,citecolor=violet!70!white,urlcolor=blue]{hyperref}

\newcommand{\R}{\mathbb{R}}
\newcommand{\E}{\mathbb{E}}
\newcommand{\argmax}{\mathop{\mathrm{argmax}}}

\newcommand{\vb}[1]{\boldsymbol{#1}}

\title{In-Context Learning for Latent Space Bayesian Optimization}

\begin{document}

\twocolumn[
\icmltitle{In-Context Learning for Latent Space Bayesian Optimization}

\begin{icmlauthorlist}
\icmlauthor{Tuan A. Vu}{1}
\icmlauthor{Harri Lähdesmäki}{1}
\icmlauthor{Julien Martinelli}{1}
\end{icmlauthorlist}

\icmlaffiliation{1}{Aalto University}
\icmlcorrespondingauthor{Tuan A. Vu}{tuan.t.vu@aalto.fi}

\vskip 0.3in
]

\printAffiliationsAndNotice{}

\begin{abstract}
Bayesian optimization (BO) is a central tool for sample-efficient design, and latent-space Bayesian optimization (LSBO) extends it to structured objects such as molecules and proteins. In parallel, tabular foundation models such as TabPFN and TabICL now achieve state-of-the-art regression performance and are increasingly used as BO surrogates. Because their Bayesian behavior is induced by large synthetic pretraining collections, the composition of this pretraining distribution is crucial. LSBO creates a distinctive mismatch: the induced map from latent code to objective value differs markedly from the regression tasks used to train current in-context models.
We address this mismatch by complementing the pretraining stage of tabular foundation model surrogates with synthetic optimization tasks defined on the latent space of a molecular VAE.
The continued-pretraining objective features a regularizer that anchors the model to the original checkpoint, preserving its broad regression prior while avoiding overspecialization to the adaptation tasks.
On held-out molecular optimization benchmarks, the resulting model achieves strong performance, supporting the relevance of LSBO-specific adaptation for in-context surrogates.
\end{abstract}

\section{Introduction}

Many scientific design problems require optimizing expensive black-box objectives over large structured spaces. Examples include molecular design, protein engineering, and sequence optimization, where each candidate may require a simulator, wet-lab assay, or human-in-the-loop evaluation \citep{gomez2018automatic,stanton2022lambo,gruver2024lambo2}. Bayesian optimization (BO) is well suited to this setting because it uses a probabilistic surrogate to balance exploration and exploitation under a limited evaluation budget \citep{garnett2023bayesian}. The main difficulty is that many design spaces are combinatorial and highly structured.

Latent-space Bayesian optimization (LSBO) addresses this by learning a continuous representation of the structured design space and performing BO in that latent space before decoding candidates back to the original domain \citep{gomez2018automatic,maus2022lolbo,lee2025nfbo,vu2026talbo}. This recipe has become standard in molecular design and increasingly relevant in protein design. The LSBO literature also shows that strong performance depends on more than a single global surrogate: successful methods rely on locality, trust regions, elite-biased updates, or representation adaptation to handle latent-geometry mismatch and decoder-induced errors \citep{tripp2020weighted,maus2022lolbo,chu2024invbo,lee2025nfbo,moss2025cowboys}.

In parallel, tabular foundation models (TFMs) and prior-data fitted networks (PFNs) have emerged as strong probabilistic predictors for tabular regression and classification \citep{mueller2021transformers,hollmann2025accurate,qu2026tabiclv2}. These models are trained on large synthetic task collections and learn to perform approximate Bayesian prediction in context. Recent works present PFNs as a broad paradigm for Bayesian prediction \citep{muller2025priorfitted,zhang2025onemodel}. This development is now reaching BO: PFNs4BO uses PFNs as surrogates, GIT-BO scales TFMs to high-dimensional BO~\citep{yu2025gitbo}, and in-context optimization policies are beginning to appear \citep{zhang2025tamo}.

The difficulty is that BO, and especially LSBO, places specific demands on the surrogate. Good decisions depend mainly on high-value regions of the objective and on the local neighborhoods defined by the best observations seen so far \citep{maus2022lolbo,yu2025gitbo}. For in-context models, this makes pretraining crucial: they inherit useful Bayesian behavior only when their synthetic training tasks resemble those seen at deployment. In LSBO, this resemblance often fails. The map from latent code to objective value is shaped by the decoder and by structured properties of the underlying object. Moreover, the objectives themselves are often domain-specific property scores such as logP, QED, or similarity-based criteria \citep{brown2019guacamol}, rather than the synthetic targets used in TFM pretraining \citep{qu2026tabiclv2}. Current BO uses of TFMs treat them as generic plug-in regressors, leaving this mismatch unresolved.

\textbf{Contributions.}
We address this mismatch through LSBO-specific adaptation of a pretrained tabular foundation-model surrogate. We construct synthetic objectives from a library of molecular base tasks derived from the GuacaMol database and sample contexts with a value-biased distribution that emphasizes promising regions of the VAE latent space. We use these episodes to continue pretraining a selected tabular foundation model checkpoint, with a regularizer that anchors the adapted model to the original weights and preserves its broad regression prior. We also introduce latent-space and objective-space diagnostics of the proposed synthetic distribution. The resulting model, \textsc{LilBO} (Latent In-context Learning for Bayesian Optimization), can be plugged directly into modern LSBO pipelines with local search, elite mechanisms, and periodic VAE retraining. On held-out molecular optimization tasks, \textsc{LilBO} achieves strong performance, supporting the relevance of LSBO-specific adaptation for in-context surrogates.

\section{Related work}

\textbf{Latent-space Bayesian optimization.}
LSBO extends BO to structured domains by introducing a learned continuous representation in which candidate generation and acquisition optimization become tractable \citep{gomez2018automatic}. Much of the recent literature focuses on the mismatch between latent geometry and the objective. Weighted retraining changes the latent data distribution toward promising regions \citep{tripp2020weighted}; LOLBO combines latent BO with local trust-region search \citep{maus2022lolbo}; CoBO and InvBO improve local alignment or invert observations into better latent codes \citep{lee2023cobo,chu2024invbo}; NF-BO uses normalizing flows to mitigate reconstruction-induced value discrepancy \citep{lee2025nfbo}; and COWBOYS argues for a stronger decoupling between the generative backbone and the surrogate \citep{moss2025cowboys}. Beyond LSBO,  agentic and language-based optimizers are also reshaping the biological design landscape \citep{maus2026pablo}.

\textbf{In-context learning for BO.}
PFNs learn posterior prediction from synthetic task distributions in a single forward pass \citep{mueller2021transformers,hollmann2025accurate,qu2026tabiclv2,grinsztajn2026tabpfn3technicalreport}. PFNs4BO showed that such models can serve as BO surrogates \citep{mueller2023pfns4bo}, while $\alpha$-PFN moved toward amortizing information-theoretic acquisition functions \citep{viering2025alphapfn}. GIT-BO showed that TFM surrogates can scale to hundreds of dimensions when coupled with active-subspace search \citep{yu2025gitbo}. More recently, fully amortized policies suggest that surrogate fitting and acquisition design may eventually be replaced end to end in some BO settings \citep{maraval2023end, zhang2025pabbo, zhang2025tamo, blumer2026context}. These results motivate our focus on LSBO, where the pretraining distribution becomes especially important.

\section{Problem statement}

Let $\mathcal{X}$ denote a structured design space, such as molecules or protein sequences, and let $f:\mathcal{X}\to\R$ be an expensive black-box objective.
We consider LSBO with a learned encoder-decoder representation, given by an encoder $e_\psi:\mathcal{X}\to\mathcal{Z}$ and a decoder $p_\psi(\vb x\mid \vb z)$ over a latent space $\mathcal{Z}\subset\R^d$. LSBO operates by searching over latent codes rather than directly over structured objects. When the decoder is stochastic, the latent optimization problem is
\begin{equation}
\vb z^\star \in \argmax_{\vb z \in \mathcal{Z}} g(\vb z),
\qquad
g(\vb z) := \E_{\vb x \sim p_\psi(\cdot \mid \vb z)}[f(\vb x)].
\end{equation}
In the deterministic case, this reduces to $g(\vb z)=f(h_\psi(\vb z))$, where $h_\psi$ is the decoder map.
Given the history
\begin{equation}
\mathcal{H}_t = \{(\vb z_i,y_i)\}_{i=1}^{t},
~~
\vb z_i = e_\psi(\vb x_i),
~~
y_i = f(\vb x_i) + \varepsilon_i,
\end{equation}
LSBO fits a probabilistic surrogate $q_t(y\mid \vb z,\mathcal{H}_t)$ and uses it to define an acquisition rule $\alpha_t$, for instance expected improvement or Thompson sampling~\cite{garnett2023bayesian}. The next latent query is selected as
\begin{equation}
\vb z_{t+1} \in \argmax_{\vb z \in \mathcal{Z}} \alpha_t(\vb z),
\end{equation}
and is then decoded and evaluated:
\begin{equation}
\vb x_{t+1} \sim p_\psi(\cdot\mid \vb z_{t+1}),
\qquad
y_{t+1}=f(\vb x_{t+1})+\varepsilon_{t+1}.
\end{equation}
Classically, the surrogate $q_t$ is taken to be the posterior predictive distribution of a Gaussian process fitted on $\mathcal{H}_t$~\citep{rasmussen2006gp}. In practice, this surrogate is often embedded in a broader pipeline that includes trust-region heuristics and retraining of the generative model.

Our goal is to replace $q_t$ with a pretrained in-context model
\begin{equation}
q_\theta(y \mid \vb z,\mathcal{H}_t),
\end{equation}
used as the predictive model inside the LSBO loop.
The parameters $\theta$ are learned offline through pretraining on a distribution of synthetic regression tasks. At test time, given the observed history $\mathcal{H}_t$ and a set of query latent points, the model outputs predictive posterior quantities for those queries. In analogy with Bayesian inference, the synthetic pretraining distribution plays the role of a flexible implicit prior over regression problems. The key question is therefore how to choose this distribution so that the resulting predictions are adapted to LSBO.

\section{Method}

Our approach has two components: a BO-aware synthetic adaptation distribution on the latent space of a pretrained molecular VAE, and a continued-pretrained TabPFN-3 surrogate used inside a standard LSBO loop.

\subsection{BO-aware synthetic task generation}
\label{sec:synthetic-task-generation}

We start from the GuacaMol molecular corpus $\mathcal{M}=\{\vb x_m\}_{m=1}^{M}$ encoded by a pretrained SELFIES VAE, yielding latent codes $\vb z_m=e_\psi(\vb x_m)$ \citep{brown2019guacamol,maus2022lolbo}.

\textbf{Synthetic target property creation.}
Rather than drawing synthetic objectives from generic regression priors, we construct them from a small library of molecular \emph{base tasks}, such as logP, QED, similarity, and rediscovery. For each synthetic episode, we sample a small subset of these tasks and evaluate them on the molecular pool. Writing
\begin{equation}
\vb s(\vb x)=\bigl(s_1(\vb x),\ldots,s_K(\vb x)\bigr)\in\R^K
\label{eq:descriptor_map}
\end{equation}
for the resulting vector of selected base-task values, we then construct a synthetic scalar objective by applying either a linear or nonlinear combiner:
\begin{align}
f_{\tau}^{\mathrm{lin}}(\vb x)
&=
\vb w_{\tau}^{\top}\vb s_\tau(\vb x),
\qquad
\vb w_\tau \sim \mathrm{Dirichlet}(\mathbf{1}_K),
\label{eq:linear-synthetic-objective}
\\
f_{\tau}^{\mathrm{nlin}}(\vb x)
&=
g_\tau\!\big(\vb s_\tau(\vb x)\big),
\label{eq:nonlinear-synthetic-objective}
\end{align}
where \(g_\tau:\R^K\to\R\) is a randomly sampled MLP. This follows the general philosophy of synthetic task generation for TFMs while grounding the prior in molecular objectives that are closer to downstream LSBO than generic synthetic targets. Additional combiner families, including formula trees, and the full base-task library are described in Appendix~\ref{app:pretraining-details}. Importantly, the evaluation benchmarks are excluded from the base-task library, so downstream objectives do not appear in the pretraining distribution.


\textbf{Context dataset sampler.}
The context sampler is designed to reflect BO rather than i.i.d.\ regression. Uniformly sampled contexts contain mostly mediocre molecules and provide limited information about high-value objective regions. We therefore bias sampling toward promising points.
Given a synthetic objective \(f_\tau\), we evaluate it on the GuacaMol pool $\mathcal{M}$ and induce a Boltzmann distribution
\begin{equation}
p_{\tau}(m)
=
\frac{\exp\!\big(f_{\tau}(\vb x_m)/T_\tau\big)}
{\sum_j \exp\!\big(f_{\tau}(\vb x_j)/T_\tau\big)},
\label{eq:sampler-round3}
\end{equation}
where \(T_\tau\) is sampled per episode (Appendix~\ref{app:context-query-sampling}). This biases contexts toward high-value regions while retaining diversity across tasks. 



\begin{algorithm}[h!]
\caption{Synthetic BO episode generation for \textsc{LilBO}}
\label{alg:pretrain-round3}
\begin{algorithmic}[1]
\REQUIRE Molecular pool $\mathcal{M}=\{\vb x_m\}_{m=1}^M$, fixed encoder $e_\psi$, descriptor map $\vb s$ in Eq.~\eqref{eq:descriptor_map}
\STATE Encode the pool once: $\vb z_m \gets e_\psi(\vb x_m) ~\forall \vb x_m\in\mathcal M$
\STATE Sample a synthetic objective $f_\tau$ from Eqs.~\eqref{eq:linear-synthetic-objective}--\eqref{eq:nonlinear-synthetic-objective}
\STATE Evaluate pool labels: $y_m^\tau \gets f_\tau(\vb x_m)$
\STATE Sample indices
$I_\tau \sim p_\tau^{\otimes\,n_{\mathrm{ctx}}+n_{\mathrm{qry}}}
$
via Eq.~\eqref{eq:sampler-round3}, and split
$
I_\tau = I_\tau^{\mathrm{ctx}} \cup I_\tau^{\mathrm{qry}},
\qquad
|I_\tau^{\mathrm{ctx}}| = n_{\mathrm{ctx}},
\quad
|I_\tau^{\mathrm{qry}}| = n_{\mathrm{qry}}.
$
\STATE Form the context and query sets

$
\mathcal{C}_\tau
 \hspace{-.05cm}= \hspace{-.05cm}
\{(\vb z_i,y_i^\tau): i\in I_\tau^{\mathrm{ctx}}\},
\mathcal{Q}_\tau
 \hspace{-.05cm}= \hspace{-.05cm}
\{(\vb z_i,y_i^\tau): i\in I_\tau^{\mathrm{qry}}\}.$
\vspace{-.3cm}
\STATE Emit the training episode
$
\mathcal{E}_\tau = (\mathcal{C}_\tau,\mathcal{Q}_\tau).
$
\STATE Update $q_\theta$ with the TabPFN-3 predictive loss and $L_2$-SP regularization toward $\theta_0$ via Eq.~\eqref{eq:loss}.
\end{algorithmic}
\end{algorithm}

\subsection{LSBO-continued-pretrained TabPFN-3 surrogate}
\label{sec:plugin-tabicl}

Rather than pre-training from scratch using our proposed molecular synthetic prior, our plug-in surrogate is initialized from a pretrained TabPFN-3 checkpoint~\citep{grinsztajn2026tabpfn3technicalreport}. The rationale is that molecular LSBO episodes are highly specialized: they involve a complex mapping involving VAE latent codes and objective ladnscapes induced by molecular property scores, and value-biased contexts. Training only on such episodes might give full control over the molecular distribution, but could also discard the broad regression prior already learned by large-scale TFM pretraining. Continued pretraining provides a compromise: the model can adapt to the geometry and target distributions of molecular LSBO while retaining its general-purpose notion of regression tasks. This reasoning is akin to that of Real-TabPFN \citep{garg2025real}, which shows that further pretraining can improve tabular foundation models on downstream data distributions. We operationalize this idea with their continued-pretraining objective. 
Let $\theta_0$ denote the original checkpoint. The continued-pretraining objective is
\begin{equation}
\mathcal{L}(\theta)
=
\mathcal{L}_{\mathrm{CE}}(\theta)
+
\frac{\lambda_{\mathrm{L2-SP}}}{2}
\|\theta-\theta_0\|_2^2,
\label{eq:loss}
\end{equation}
where the second term limits drift from the pretrained model and mitigates catastrophic forgetting~\citep{kirkpatrick2017overcoming}. Full optimization details are given in Appendix~\ref{sec:training}.
Algorithm~\ref{alg:pretrain-round3} summarizes the training loop.
We train for 8,000 continued-pretraining steps, exposing the model to approximately 500k molecular LSBO episodes. Full optimization details are given in Table~\ref{table:con_pretrained_setup}.


At test time we keep the LSBO loop unchanged and replace only the surrogate. For each candidate batch $\mathcal{Z}^{\mathrm{cand}}_t$, typically produced by local search, we retrieve up to $n_{\mathrm{ctx}}$ nearby or high-value observations from the BO history and score the candidates with the pretrained model. The next query is then selected by Thompson sampling,
\begin{equation}
\widetilde f_t(\vb z) \sim q_\theta(\cdot\mid \vb z,\widetilde{\mathcal{H}}_t),
\qquad
\vb z_{t+1} = \argmax_{\vb z\in\mathcal{Z}^{\mathrm{cand}}_t} \widetilde f_t(\vb z).
\end{equation}
In the current instantiation,  continued-pretraining is performed on the latent space of a frozen VAE. At test time, however, we keep the LSBO pipeline unchanged and allow periodic VAE retraining, since this yields substantial benefits~\citep{maus2022lolbo}. Hence, the latent space drifts during optimization, creating a mismatch between the implicit prior learned during pretraining and the tasks encountered at deployment. This is a current limitation of our approach.
\section{Experiments}

\textbf{Benchmarks and protocol.}
We use the SELFIES VAE of \citet{maus2022lolbo}, pretrained on the 1.27M GuacaMol molecules \citep{brown2019guacamol}, as the base generator for synthetic pretraining. Crucially, downstream evaluation is performed on held-out GuacaMol benchmark objectives that were not seen during pretraining, thereby testing generalization to unseen objectives. Our main suite contains eight molecular multi-property optimization tasks: Osimertinib, Fexofenadine, Ranolazine, Zaleplon, Amlodipine, Perindopril, Median Molecules 1 and Median Molecules 2 \citep{brown2019guacamol}. Each run starts from an initial set of 100 randomly sampled molecules, and we report the best value found versus oracle calls over 10 random seeds. The setup follows \cite{lee2023cobo} and is designed to emulate real-world lead optimization.

\begin{figure*}[h]
\centering
\includegraphics[width=\linewidth]{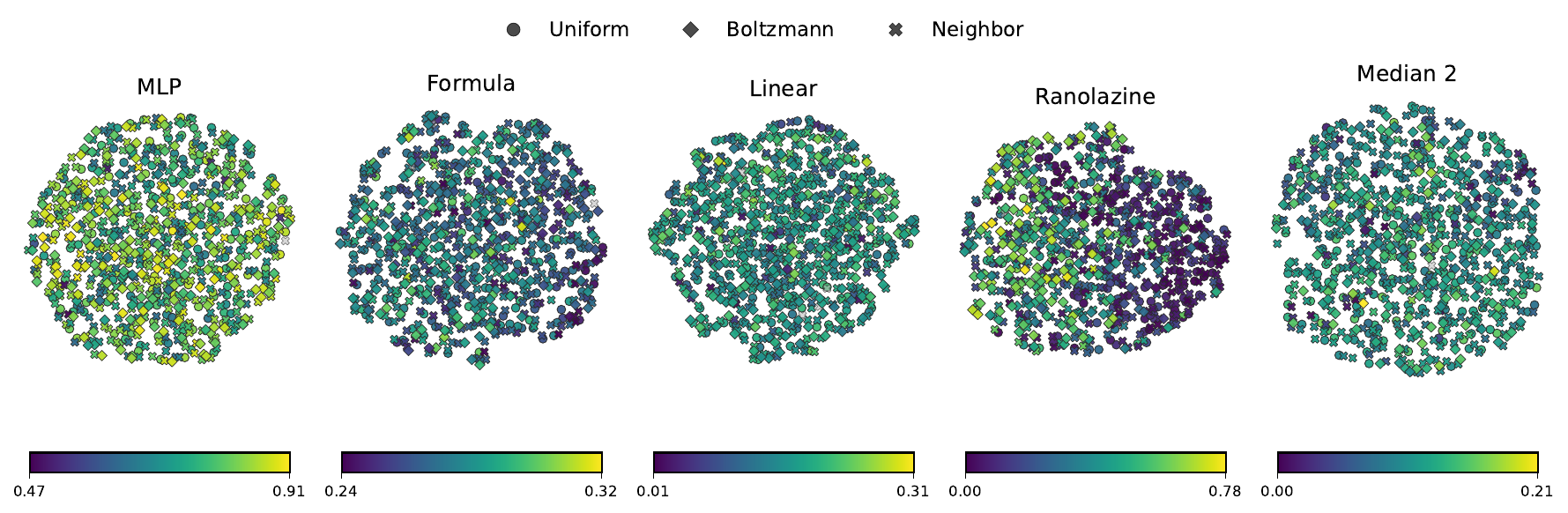}
\caption{
\textbf{Latent-space coverage of pretraining episodes.}
UMAP embeddings for 3 synthetic objectives and 2 held-out benchmarks.
}
\label{fig:latent-prior-umap}
\end{figure*}

\begin{figure}[h]
\centering
\includegraphics[width=\linewidth]{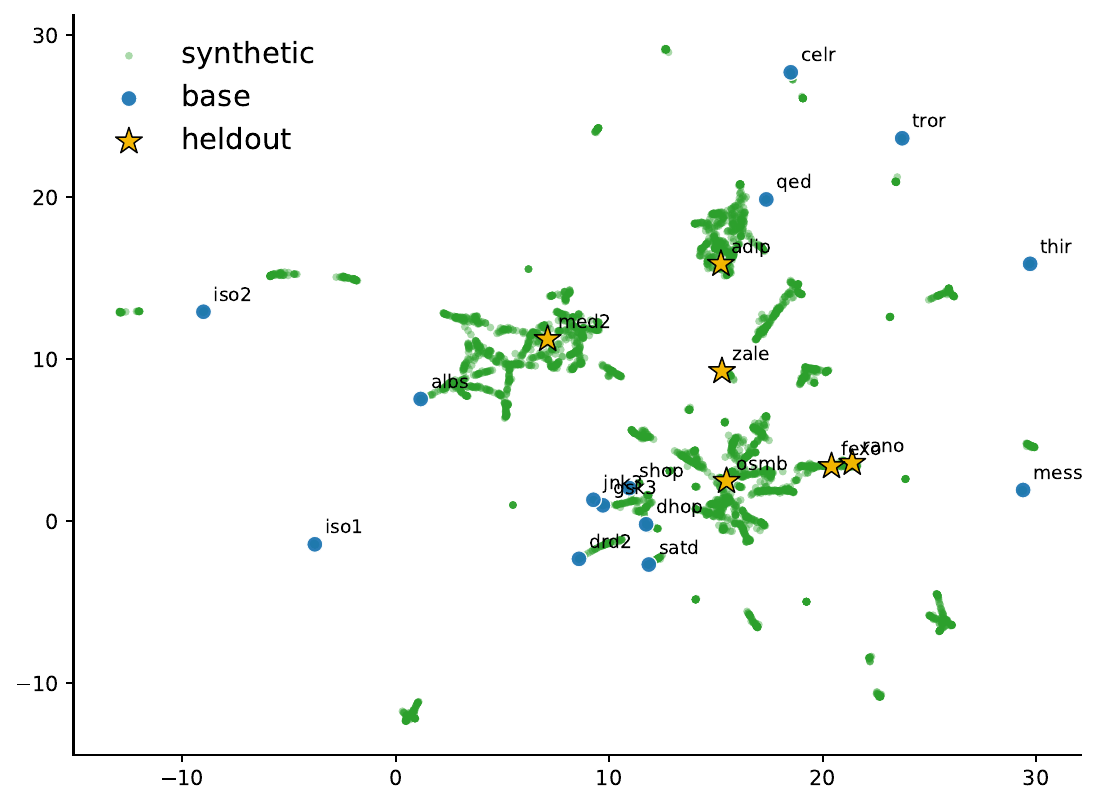}
\caption{
\textbf{Objective-space coverage of the synthetic prior.}
Each objective is evaluated on the same probe set $\mathcal{P}$ and represented by the resulting value vector (Equation~\ref{eq:objective-vector}), embedded with UMAP.
}
\label{fig:objective-prior-umap}
\end{figure}

\textbf{Baselines and implementation details.}
We compare against representative LSBO baselines, including GP-LSBO~\citep{eriksson2019turbo, gomez2018automatic}, LOLBO~\citep{maus2022lolbo}, CoBO~\citep{lee2023cobo}, InvBO~\citep{chu2024invbo}, NF-BO~\citep{lee2025nfbo} and random latent search.
To isolate the effect of LSBO-specific continued pretraining, we also include an off-the-shelf TabPFN-3 plug-in baseline~\citep{grinsztajn2026tabpfn3technicalreport}. The latter uses the same BO loop and candidate batches as our method, but without our molecular LSBO adaptation.
All baselines select the next query via Thompson sampling over a common candidate set $\mathcal{Z}^{\mathrm{cand}}_t$, whose generation is detailed in Appendix~\ref{app:candidate-sampling}. Moreover, all baselines except LSBO and random perform VAE retraining. For 
non-TFM baselines this entails joint retraining of the surrogate and the VAE, whereas our TFM-based methods replace only the surrogate. This tests whether a pretrained in-context model remains effective under the latent drift induced by strong LSBO pipelines.

\begin{figure*}[h!]
\centering
\includegraphics[width=1\textwidth]{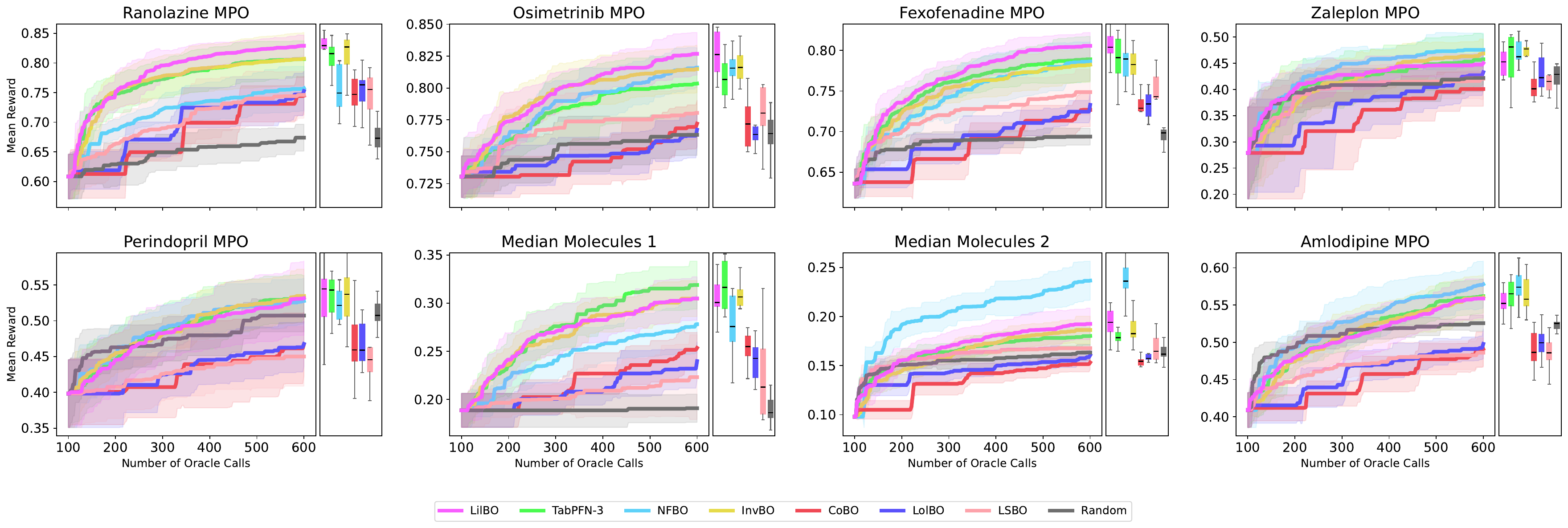}
\caption{\textbf{Main results.} Best value found over oracle calls and distribution over the final recommendation on held-out molecular objectives. Mean $\pm$ 1 std computed across 10 seeds.}
\label{fig:main-results}
\end{figure*}

\subsection{Pretraining-prior diagnostics}

Before turning to BO performance, we introduce two diagnostics of the synthetic adaptation distribution, designed to assess whether it covers the latent-space regions and objective-space variability relevant to LSBO.

\textbf{Latent-space coverage of sampled episodes.}
For a fixed objective $f_\tau$, we define three sets of latent points:
\begin{enumerate}[leftmargin=1.2em]
    \item \emph{Uniform pool samples:}
    $
    \mathcal{Z}_{\tau}^{\mathrm{uni}}
    =
    \{\vb z_l\}_{l=1}^{L},
    $
    where the corresponding molecules are sampled uniformly from the GuacaMol pool $\mathcal{M}$.

    \item \emph{Value-biased pool samples:}
    $
    \mathcal{Z}_{\tau}^{\mathrm{bias}}
    =
    \{\vb z_l\}_{l=1}^{L},
    $
    where the corresponding molecules are sampled from the same pool $\mathcal{M}$ with the value-biased distribution in Equation~\ref{eq:sampler-round3}.

\item \emph{Local perturbations:}
$
\mathcal{Z}_{\tau}^{\mathrm{pert}}
=
\left\{
\vb z_i + r \vb \xi_i :
\vb z_i \in \mathcal{Z}_{\tau}^{\mathrm{uni}} \cup \mathcal{Z}_{\tau}^{\mathrm{bias}}
\right\}$, with $\vb \xi_i \stackrel{\mathrm{i.i.d.}}{\sim} \mathrm{Unif}(\mathbb{B}_d)$ and $r$ small.
\end{enumerate}
We fit a two-dimensional UMAP to $\mathcal{Z}_{\tau}^{\mathrm{all}} := \mathcal{Z}_{\tau}^{\mathrm{uni}} \cup \mathcal{Z}_{\tau}^{\mathrm{bias}} \cup \mathcal{Z}_{\tau}^{\mathrm{pert}} $ and display a single embedding in which marker shape indicates provenance and color indicates objective value. This visualization checks whether value-biased sampling and local perturbations populate latent regions that are relevant to BO, and whether these regions align with high-value parts of the landscape.

Figure~\ref{fig:latent-prior-umap} shows the representation for $L=200$, so that $800$ points are displayed.
The main question is whether value-biased samples and their local perturbations are enriched in higher-value regions relative to uniform samples. In the current plots, the objective-value coloring does suggest that the landscapes differ across tasks, with some objectives exhibiting more localized high-value regions than others. At the same time, provenance differences remain moderate, which indicates that the perturbation mechanism stays close to the latent manifold. This also means that the figure alone does not fully characterize the gap between pretraining episodes and test-time BO candidate distributions.

\textbf{Objective-space coverage of synthetic tasks.}

We also visualize the diversity of the synthetic objective family. Let $\mathcal{P}=\{\vb x_j\}_{j=1}^{J}$ be a fixed molecular probe set. It is not necessary that $\mathcal{P} \subset \mathcal{M}$. Each objective $f_\tau$ is represented by its vector of values on this probe set,
\begin{equation}
\vb v_\tau
=
\big(
f_\tau(\vb x_1),\ldots,f_\tau(\vb x_J)
\big)
\in [0,1]^m.
\label{eq:objective-vector}
\end{equation}
We compute these vectors for both synthetic pretraining objectives and held-out benchmark objectives, then embed them in two dimensions with UMAP. The comparison is inherently probe-set dependent and should be interpreted carefully. Still, it shows whether the synthetic prior covers a broad region of molecular objective space.

Figure~\ref{fig:objective-prior-umap} displays the representation for 
$|\mathcal{P}| = 5000$ and $5000$ synthetic tasks, as well as the held-out and base tasks.
The synthetic tasks form continuous clusters around the base objectives, as expected from linear and nonlinear combinations of molecular descriptors. Importantly, the held-out GuacaMol benchmarks do not appear isolated, lying close to the synthetic task mass. This supports the intended role of the prior: exposing the model to related molecular optimization landscapes. For completeness, base objective distributions and representative synthetic objective distributions are shown in Figures~\ref{fig:base-diagnostics-round3} and~\ref{fig:synthetic-diagnostics-large}.

\subsection{Results}

We next evaluate whether our proposed pretraining distribution improves downstream optimization on held-out molecular objectives. Figure~\ref{fig:main-results} reports the best value found as a function of oracle evaluations, and Table~\ref{tab:ranks} shows baseline ranking aggregated over tasks and seeds. 

\textbf{Tabular foundation models are strong plug-in surrogates for LSBO.}
Figure~\ref{fig:main-results} shows that tabular foundation models can already serve as competitive surrogates for LSBO: both off-the-shelf TabPFN-3 and \textsc{LilBO} are in the leading group across the held-out molecular objectives. Importantly, the optimization pipeline is otherwise kept fixed, including candidate generation, acquisition rule, and VAE retraining schedule, so the comparison isolates the surrogate replacement itself. These results suggest that in-context probabilistic surrogates are a viable alternative to GPs in modern LSBO pipelines.

\begin{table}[h]
\centering
\begin{tabular}{lc}
\hline
Method & Average rank \\
\hline
\textsc{LilBO} & $2.64 \pm 1.56$ \\
InvBO & $2.88 \pm 1.62$ \\
TabPFN-3 & $2.94 \pm 1.64$ \\
NFBO & $2.96 \pm 1.69$ \\
LSBO & $5.86 \pm 1.75$ \\
LolBO & $5.90 \pm 1.37$ \\
Random & $6.25 \pm 1.68$ \\
CoBO & $6.34 \pm 1.19$ \\
\hline
\end{tabular}
\caption{\textbf{Average rank of the methods across all tasks.} Ranks were computed separately for each seed, the table reports the mean rank $\pm$ 1 std across seeds.}
\label{tab:ranks}
\end{table}

\textbf{LSBO-aware continued pretraining improves average performance against TabPFN-3.}
Table~\ref{tab:ranks} shows that \textsc{LilBO} slightly improves over the TabPFN-3 plug-in baseline overall, with average ranks of $2.64$ and $2.94$, respectively. This gain is modest but notable, as our adaptation is intentionally lightweight compared with the original TabPFN-3 pretraining. The checkpoint was trained on more than 8 trillion synthetic tokens~\citep{grinsztajn2026tabpfn3technicalreport}, whereas our molecular adaptation uses only about 500k LSBO episodes of length 1024.
On a similar note, TabICLv2 was pretrained for 550K synthetic-data steps with batch size 64, corresponding to roughly 35M synthetic datasets~\citep{qu2026tabiclv2}.
Moreover, our adaptation distribution is far more specialized, operating on 256-dimensional SELFIES-VAE latents with 100--600 context rows, compared with the broad row--feature regimes targeted by TabPFN-3.
Overall, this motivates anchoring the adapted model to the original checkpoint: the goal is to specialize the surrogate to molecular LSBO while preserving the broad regression prior learned during large-scale pretraining.

\section{Conclusion}

We studied whether in-context surrogates can be adapted to latent-space Bayesian optimization through task-matched pretraining. Our main empirical message is that tabular foundation models are already competitive surrogates for LSBO when inserted into an otherwise fixed optimization pipeline, and that LSBO-aware continued pretraining yields additional gains. While \textsc{LilBO} does not uniformly dominate every task or every specialized LSBO baseline, it achieves the best average rank in our benchmark suite, supporting the central hypothesis of this work:
For tabular foundation models to be effective in LSBO, their adaptation distribution should reduce the mismatch between generic tabular pretraining tasks and the structure of latent space objectives and BO histories encountered at deployment.


\textbf{Perspectives.}
Several limitations suggest natural next steps. First, pretraining episodes should be made more BO-like by simulating short optimization trajectories rather than sampling only from the molecular pool, so that training contexts include high-value molecules beyond those already present in GuacaMol. This would better match deployment histories and is consistent with recent work \citep{li2025profbo}.
Second, the prior should account for the latent drift induced by periodic VAE retraining, so that the in-context surrogate is pretrained not only on synthetic objectives but also on evolving latent representations. Third, multi-objective LSBO is a natural extension, particularly in drug design where one must balance properties such as potency, toxicity, and solubility. Finally, the same framework should extend beyond small-molecule design to other structured domains, including protein and antimicrobial peptide design, provided a suitable generative model supplies a latent representation.
\bibliographystyle{plainnat}
\bibliography{tfm_lsbo_refs}

\clearpage
\appendix

\onecolumn

\section{Synthetic pretraining distribution}
\label{app:pretraining-details}

\subsection{Descriptor library and objective families}
\label{app:descriptor-objectives}


To construct a synthetic objective, we first draw
$K\sim\mathcal{U}(\{1,2, 3,4\}),
$
and then sample \(K\) base objectives from Table~\ref{table:prior_tasks}. 
\begin{table}[h!]
\begin{center}
\begin{tabular}{cll}
\hline
Index & Objective function \\
\hline
1 & albuterol\_similarity \\
2 & mestranol\_similarity \\
3 & celecoxib\_rediscovery \\
4 & thiothixene\_rediscovery \\
5 & troglitazone\_rediscovery \\
6 & isomer\_c7h8n2o2 \\
7 & isomer\_c9h10n2o2pf2cl \\
8 & sa\_tdc \\
9 & drd2\_docking \\
10 & rdkit\_qed \\
11 & deco\_hop \\
12 & scaffold\_hop \\
13 & gsk3\_beta \\
14 & jnk3 \\
\hline
\end{tabular}
\end{center}
\caption{Descriptors used for synthetic prior tasks.}
\label{table:prior_tasks}
\end{table}

For a molecule \(\vb x\), let $\vb{s}(\vb x)=\bigl(s_1(\vb x),\ldots,s_K(\vb x)\bigr)$
denote the selected objective values. We then combine them using a
linear-weighted combiner, a random MLP, or a formula tree.

\paragraph{Linear combiner.}
The linear-weighted family combines the selected molecular objectives through a
random weighted average. We draw the weights from a symmetric Dirichlet
distribution,
\[
\mathbf{w}\sim\mathrm{Dirichlet}(\mathbf{1}_K),\qquad
f_{\text{lin}}(\vb x)
=\sum_{i=1}^{K}w_i s_i(\vb x).
\]


\paragraph{Random MLP combiner.}
The random MLP family introduces smooth nonlinear interactions between the
selected molecular objectives by applying a sampled function
\(
g_\tau:\R^K\to\R
\)
to the vector \(\vb s(\vb x)\). Concretely,
\[
f_{\mathrm{mlp}}(\vb x)=g_\tau\!\big(\vb s(\vb x)\big),
\]
where \(g_\tau\) is an \(L\)-hidden-layer MLP with width \(H\), activation
\(\phi\), and block-sparse Gaussian weights. We sample
\[
L\in\{1,2,3\},\qquad
H\in\{16,32,64\},\qquad
\phi\in\{\tanh,\,\mathrm{ReLU},\,\mathrm{ELU},\,\sigma\}.
\]
Writing \(\vb h^{(0)}=\vb s(\vb x)\), the hidden states satisfy
\[
\vb h^{(\ell)}
=
\phi_\ell\!\left(
W_\ell \vb h^{(\ell-1)}+\vb b_\ell
\right),
\qquad \ell=1,\dots,L,
\]
and the final output is
\[
f_{\mathrm{mlp}}(\vb x)=
g_\tau\!\big(\vb s(\vb x)\big)
=
\vb w_{\mathrm{out}}^\top \vb h^{(L)}+b_{\mathrm{out}}.
\]

\paragraph{Formula-tree combiner.}
The formula-tree family introduces compositional objectives with gates, sharp
thresholds, and multiplicative interactions. We sample a random binary
expression tree \(f_{\mathrm{tree}}\) whose leaves are indexed by the selected objectives
\(s_1,\dots,s_K\). Internal nodes are assigned binary operations
\[
\mathcal{B}=\{+,\times,\mathrm{gate}\},\qquad
\mathrm{gate}(a,b)=\mathbf{1}[a>0]\cdot b,
\]
and unary operations
\[
\mathcal{U}=\{\mathrm{id},\,|\cdot|,\,\sin,\,\sigma,\,\mathrm{ReLU},\,
\sqrt{\cdot\,}\}.
\]
Writing \(F_v(\vb s)\) for the value at node \(v\), we define the tree
recursively as
\[
F_v(\vb s)=
\begin{cases}
s_i, & \text{if } v \text{ is a leaf labeled } i,\\[0.4em]
u_v\!\Big(
b_v\big(F_{v_{\mathrm{left}}}(\vb s),\,F_{v_{\mathrm{right}}}(\vb s)\big)
\Big), & \text{if } v \text{ is an internal node},
\end{cases}
\]
where \(b_v\in\mathcal{B}\) and \(u_v\in\mathcal{U}\). The final objective is
then
\[
f_{\mathrm{tree}}(\vb x)=F_{\mathrm{root}}(\vb s(\vb x)).
\]


\begin{figure*}[h]
\centering
\includegraphics[width=0.96\textwidth]{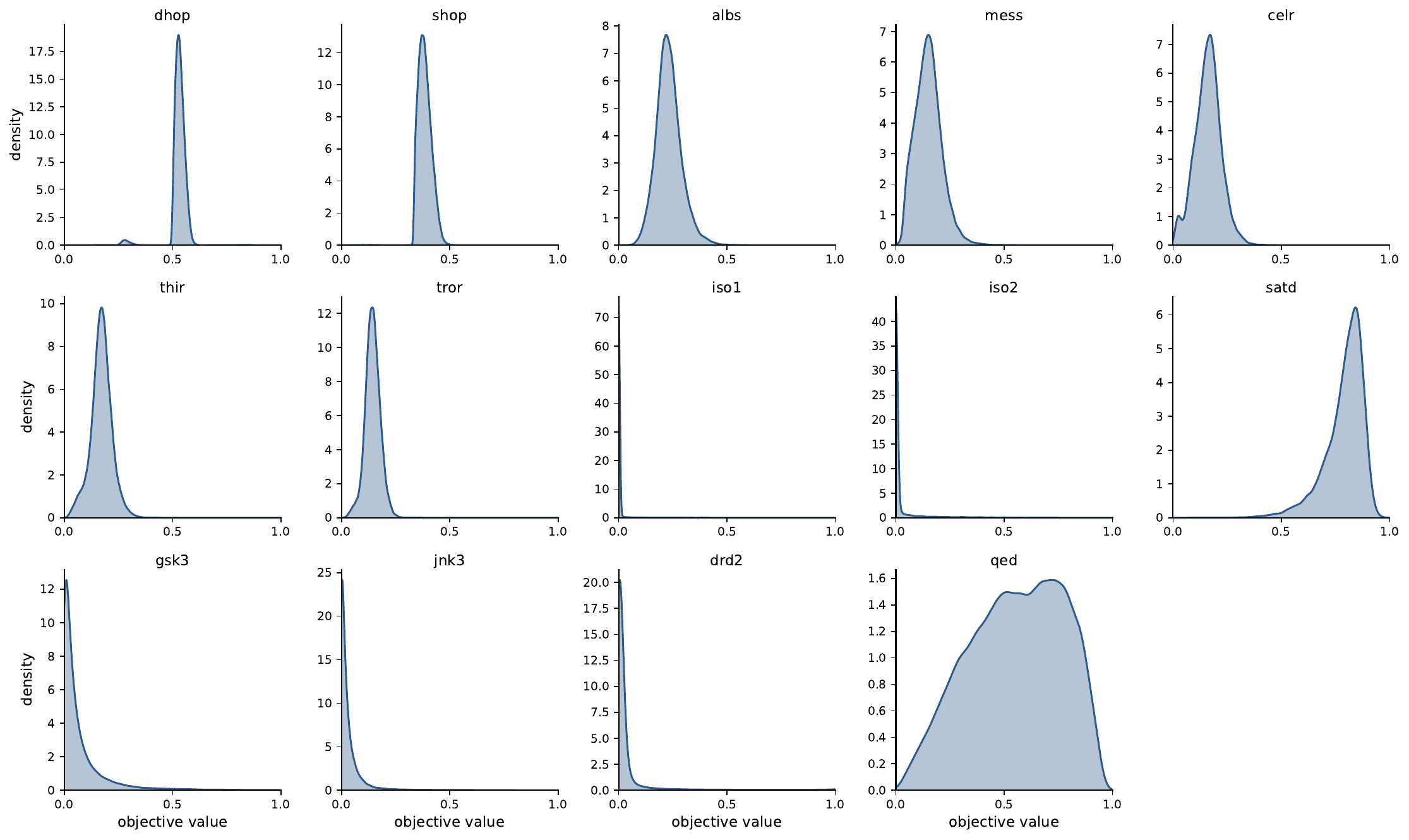}
\caption{\textbf{Base objective distributions.} Kernel density estimates of the normalized base objectives used to construct the synthetic prior.}
\label{fig:base-diagnostics-round3}
\end{figure*}

\begin{figure*}[h]
\centering
\includegraphics[width=0.96\textwidth]{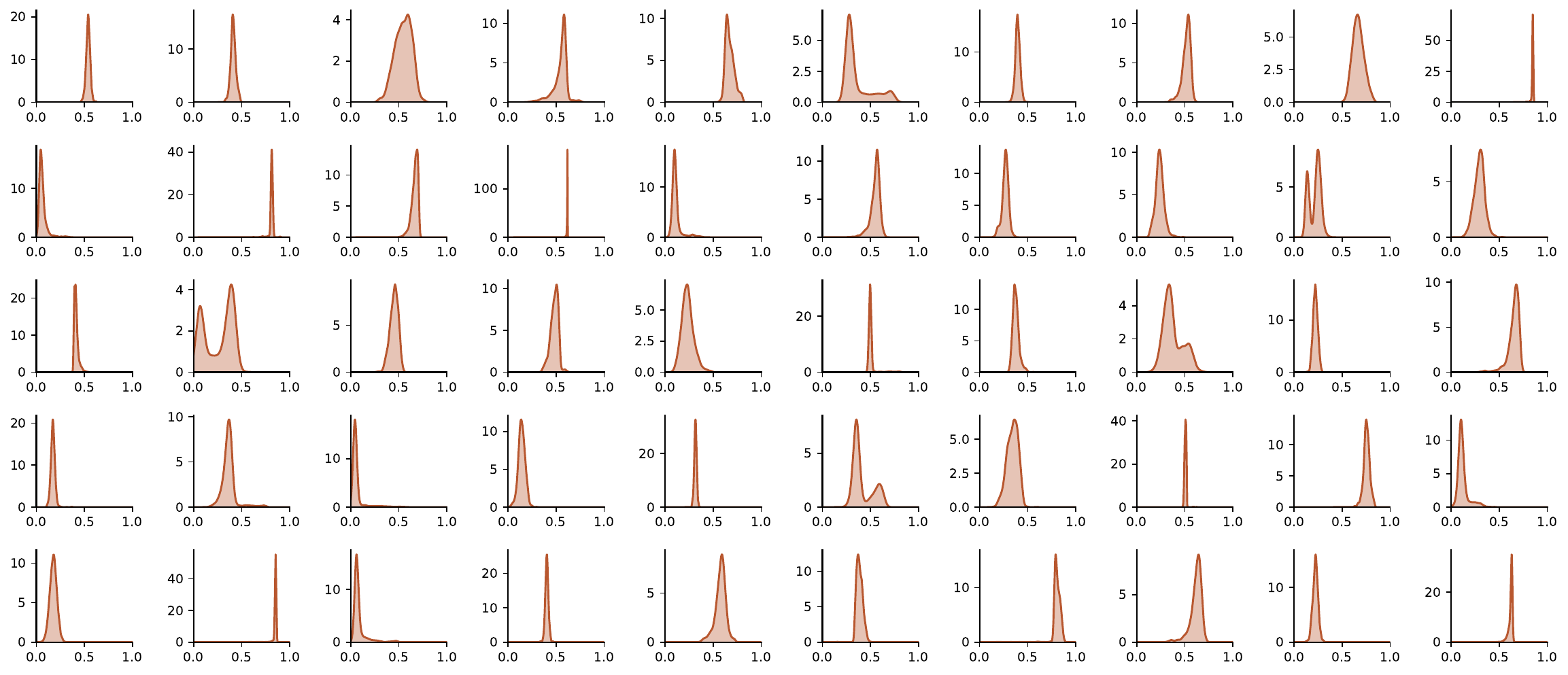}
\caption{\textbf{Synthetic objective distributions.} Kernel density estimates of representative normalized synthetic objectives sampled from the pretraining prior.}
\label{fig:synthetic-diagnostics-large}
\end{figure*}

\subsection{Context and query sampling}
\label{app:context-query-sampling}

We evaluate the synthetic objective on the GuacaMol molecular pool to obtain
\[
y_n=f\bigl(\vb s(\vb x_n)\bigr),\qquad n=1,\ldots,N,
\]
which induces a Boltzmann distribution over the molecular pool. To generate a
synthetic dataset, we draw a temperature \(T\) log-uniformly from
\[
\log T\sim\mathrm{Uniform}(\log t_{\min},\log t_{\max}),\qquad
[t_{\min},t_{\max}]=[0.01,\,10].
\]
The induced distribution is
\[
\pi_n(T)=\frac{\exp(y_n/T)}{\sum_{m=1}^{N}\exp(y_m/T)},
\qquad n=1,\ldots,N.
\]
The resulting synthetic dataset is
\[
\mathcal{D}=\{(\vb x_{i_r},\,y_{i_r})\}_{r=1}^{m}.
\]
This dataset is then partitioned into context and target points for PFN
training.




\section{Implementation details}



\subsection{Continued-pretraining configuration}\label{sec:training}




Following \cite{garg2025real}, we adopt a two-stage approach. In the first stage, we start from the original TabPFN-3 \citep{grinsztajn2026tabpfn3technicalreport} checkpoint, which has already been pretrained on a broad collection of synthetic tabular datasets. We then continue pretraining from this checkpoint using our own synthetic datasets described earlier. We keep the same architecture as TabPFN-3 and use a learning rate of $3 \times 10^{-7}$ with the AdamW optimizer \citep{loshchilov2017decoupled}, combined with a linear warm-up phase and a subsequent cosine annealing schedule \citep{loshchilov2016sgdr}. The complete training configuration is provided in Table~\ref{table:con_pretrained_setup}. To mitigate catastrophic forgetting \citep{kirkpatrick2017overcoming}, we incorporate an L2-SP regularization term \citep{xuhong2018explicit} into the training objective, preventing the model parameters from drifting too far from the weights of the original pretrained model:

\[
\mathcal{L}
=
\mathcal{L}_{CE}
+
\frac{\lambda_{L2-SP}}{2}
\lVert \theta - \theta_0 \rVert_2^2,
\]

where $\theta_0$ denotes the parameters of the initial checkpoint and $\|\cdot\|_2$ denotes the $\ell_2$ norm. The final model was trained for 8,000 steps, during which it was exposed to approximately 500k datasets.

\begin{table}[h]
\centering
\small
\begin{tabular}{p{0.16\linewidth}p{0.36\linewidth}}
\hline
\textbf{Parameter} & \textbf{Value} \\
\hline
Base model & Pretrained TabPFN-v3 checkpoint \\
Training data & Molecular synthetic datasets \\
Input features & 256-d SELFIES-VAE latents \\
Sequence length & 1024 \\
Context rows & 100--600 \\
Optimizer & AdamW \\
LR & $3 \times 10^{-7}$ \\
Weight decay & 0.01 \\
Batch size & 64 \\
Steps & 8000 \\
$\lambda_{\mathrm{L2-SP}}$ & 0.003 \\
\hline
\end{tabular}
\caption{\textbf{Training configuration used for continued pretraining.}}
\label{table:con_pretrained_setup}
\end{table}

\subsection{Baselines.}\label{sec:baselines}
We compare \textsc{LilBO} to representative latent-space optimization baselines:
\begin{itemize}[leftmargin=*]
    \item \textbf{Random search}: samples latent codes from the proposal distribution and decodes them without fitting a surrogate.

    \item \textbf{LSBO}~\citep{eriksson2019turbo}: latent-space BO with a Gaussian-process surrogate over latent codes,
    \[
    g(\vb z) \sim \mathcal{GP}(0,k(\vb z,\vb z')),
    \]
    combined with adaptive local trust regions that restrict candidate generation to a neighborhood of the current incumbent.

    \item \textbf{LOL-BO}~\citep{maus2022lolbo}: local latent-space BO with periodic retraining of the generative model, designed to improve the latent representation around promising regions.

    \item \textbf{CoBO}~\citep{lee2023cobo}: an LSBO method that learns a correlated latent space to better align latent geometry with the objective.

    \item \textbf{InvBO}~\citep{chu2024invbo}: an inversion-based LSBO method that reduces mismatch between latent points and decoded molecules.

    \item \textbf{NF-BO}~\citep{lee2025nfbo}: a latent BO method based on autoregressive normalizing flows, designed to reduce the reconstruction-gap mismatch between latent points and decoded molecules; it also uses a candidate sampling strategy that adapts exploration to token importance.

    \item \textbf{TabPFN-3 plug-in}~\citep{grinsztajn2026tabpfn3technicalreport}: replaces the GP surrogate with the released TabPFN-3 model, while using the same candidate batches, acquisition rule, and host LSBO loop as \textsc{LilBO}.
\end{itemize}

\subsection{Candidate Sampling}\label{app:candidate-sampling}

For all probabilistic surrogates, candidates are selected by Thompson sampling over the same candidate set $\mathcal{Z}^{\mathrm{cand}}_t$:
\begin{equation}
\tilde g_t \sim q_t(\cdot \mid \mathcal H_t),
\qquad
\vb z_{t+1}
\in
\argmax_{\vb z\in \mathcal{Z}^{\mathrm{cand}}_t}
\tilde g_t(\vb z).
\end{equation}
We adopt a multi-center trust-region-based local search BO method \citep{eriksson2019turbo, maus2022lolbo} in latent
space. At each BO iteration, we select \(M\) high-scoring anchor points, each of
which defines a promising local region for trust-region-based local search. A
separate trust region is constructed around each center so that the acquisition
step can explore multiple promising regions simultaneously.

A trust region centered at \(\boldsymbol{\kappa}\in\mathbb{R}^d\) with side
length \(\rho>0\) and weight \(\boldsymbol{\omega}\in\mathbb{R}^d\) is
\begin{align}
\mathcal{T}(\boldsymbol{\kappa};\rho,\boldsymbol{\omega})
&= \big\{ z\in\mathbb{R}^d : \boldsymbol{\kappa} - \tfrac{1}{2}\boldsymbol{\omega}\rho
\le z \le
\boldsymbol{\kappa} + \tfrac{1}{2}\boldsymbol{\omega}\rho \big\}.
\end{align}
Given centers \(\{\boldsymbol{\kappa}_m\}_{m=1}^{M}\), this
defines a collection of local regions
\(\{\mathcal{T}(\boldsymbol{\kappa}_m;\rho,\boldsymbol{\omega})\}_{m=1}^{M}\).

For each center, we generate the same number of candidates within its trust
region by perturbing a subset of latent dimensions. The per-center candidate
sets are pooled, and Thompson sampling from the surrogate predictive
distribution selects the final batch of latent points.





\end{document}